\documentclass[12pt]{article} 
\usepackage{times}  
\usepackage{helvet}  
\usepackage{courier}  
\usepackage{url}  
\usepackage{graphicx}  
\def\PAL{\textsc{PAL}}
\usepackage{bm}
\usepackage{mathbbol}
\usepackage{amsmath,amssymb}
\DeclareSymbolFontAlphabet{\amsmathbb}{AMSb}%
\usepackage{algorithm}
\usepackage{algorithmic}
\usepackage{xcolor}
\usepackage{tikz}
\usetikzlibrary{positioning}
\usetikzlibrary{arrows,automata}
\usepackage{pgfplots}
\pgfplotsset{compat=newest}
\DeclareMathOperator*{\argmax}{argmax}
\usetikzlibrary{arrows,automata}

\title{Incremental learning abstract discrete planning domains\\ 
  and mappings to continuous perceptions}
\author{Luciano Serafini and Paolo Traverso \\ 
Fondazione Bruno Kessler \\ Trento, Italy \\ \texttt{serafini|traverso@fbk.eu}}

\def\updgamma{\mathsf{update\_trans}}
\def\updf{\mathsf{update\_perc}}
\def\R{\mathbb{R}}

\def\N{\mathcal{N}}

\def\bmu{\pmb{\mu}}
\def\bSigma{\pmb{\Sigma}}
\def\bx{\pmb{x}}
\def\Tr{{Tr}}
\def\Obs{{Obs}}
\def\prm{\pmb{\theta}}
\newtheorem{definition}{Definition}
\newtheorem{example}{Example}

\def\PD{\mathcal{D}}
\def\PDdef{\left<S,A,\gamma\right>}
\def\PP{\mathcal{P}}
\def\dom{\pmb{Dom}}
\begin{document}
\maketitle
%

\begin{abstract}
  Most of the works on planning and learning, e.g., planning by (model
  based) reinforcement learning, are based on two main assumptions: (i)
  the set of states of the planning domain is fixed; (ii) the mapping
  between the observations from the real word and the states is
  implicitly assumed or learned offline, and it is not part of the planning domain.
  Consequently, the focus is on learning the transitions between
  states. 
  In this paper, we drop such assumptions.  We provide a formal
  framework in which (i) the agent can learn dynamically new states of
  the planning domain; (ii) the mapping between abstract states and
  the perception from the real world, represented by continuous
  variables, is part of the planning domain; (iii) such mapping is
  learned and updated along the ``life'' of the agent.  We define an
  algorithm that interleaves planning, acting, and learning, and
  allows the agent to update the planning domain depending on how much
  it trusts the model w.r.t. the new experiences learned by executing
  actions. We define a measure of coherence between the planning
  domain and the real world as perceived by the agent.
  We test our approach showing that the agent learns increasingly coherent
  models, and that the system can scale to deal with models with an
  order of $10^6$ states. 
\end{abstract}

\section{Introduction and Motivations}


Several automated planning techniques are based on abstract
representations of the world, usually called \emph{planning
  domains}. A planning domain can be formalized by a finite state
transition system,%
i.e., a finite set of states,
actions, and a transition relation \cite{ghallab2004,ghallab2016}.
This abstract representation is both conceptually relevant and
practically convenient.
Indeed, there are many domains where it is clearly convenient to plan in a discrete space.
%
For instance, in order
to plan how a robot can move packs from a room to another room in a building,
it may be convenient to adopt a planning domain where states correspond to (the fact that the robot and the packs are in) certain 
rooms, and transitions correspond to abstract actions like moving the robot between adjacent rooms,
picking up blocks, and delivering them.

%
%
%
%


While an agent can conveniently plan at the abstract level, it perceives
the world and acts in it through sensors and actuators that work with
data in a continuous space, typically represented with variables on
real numbers.
For instance, a robot does not perceive directly the fact that it is
in a given room/state, instead it perceives, e.g., to be in a position
of the building through sensors like odometers or the images
from its camera.
Similarly agent actions' effects in the
environment are continuous transformations, e.g., ``the robot has moved forward $5.4$ meters''. 
It is part of the cognitive capability of the agent to fill the gap between
these two different levels of abstractions. 

Most of the works in planning and learning, see, e.g.,
\cite{sutton98,geffner2013} assume that (i) the finite set of states
of the planning domain
is fixed once forever at design time, 
and (ii) the correspondence between the abstract states and the observations (represented with continuous variables) is implicit and fixed at design time.
This is the case of most of the works on planning by (model based)
reinforcement learning,
see, e.g.,
\cite{sutton90,sutton98,yang2018,parr97,ryan2002,leonetti2016}%
\footnote{
  In
   some works (see, e.g., \cite{abbeel2006,mnih2015,coeeyes2018}) the
   two levels are collapsed, since planning is performed in a
   continuous space.
 },
which focus on learning and updating the
transitions between states, e.g., the probabilities of
action outcomes (or rewards) in an MDP framework.  They support neither the
learning of new states corresponding to unexpected situations the agent may encounter, 
nor the updating of the mapping between the perceptions represented with continuous variables 
and the abstract discrete model.

In many cases, however, having a fixed set of states and a fixed
mapping between the perceived data and the abstract model is not
adequate.
There may be situations in which the agent perceives data which are
not compatible with any of the states of its abstract model.  For
instance, a robot may end up in unknown and unexpected states of the
world. Consider the simple example in which the task is to navigate in
a restricted part of a building, and instead, due to some reasons,
like a navigation error, or an unexpected open door, the robot ends up
in a different part of the building. 
Similarly, along its life, an
agent could also revise its mapping between its abstract model and the
real sensed data.
In general, the (number of) states and the mapping to perceptions may be not obvious at design time,
and thus be incomplete or not adequate.

%

In this paper, we provide a formal framework in which
the agent can learn dynamically new states of the planning domain.
Moreover, 
the mapping between abstract states and perceptions from the real world is part of the
planning domain of the agent, and it is learned and updated along the
``life'' of the agent. Given this framework, we provide the following contributions:
(i) We model agent's perception of the real world by a
\emph{perception function} that returns the likelihood of observing
some continuous data being in a state of the domain.
We define a criteria based on the perception function to extend
the set of states.  Intuitively, when the likelihood is too low for all
the existing states, a new state is created;
(ii) We define an algorithm that interleaves planning, acting, and
learning. It is able to discover that the abstract model is not
coherent with the real world.  While planning and acting, the
algorithm updates both the set of states and the perception function of the
planning domain;
(iii)
the learning of the planning domain can be defined through some key parameters 
that allow the agent either to follow a cautious strategy, 
where changes are made only if there is a certain number of evidences from acting
and perceiving the real world, or a more impulsive reaction to what
the agent has just observed;
(iv)
we define a measure of coherence between the planning domain and
the real world as perceived by the agent;
(v) we provide experiments that show the scalability of the approach
and also thoroughly analyze 
how setting these parameters influence the learning process and
the convergence to a model coherent with the world. 


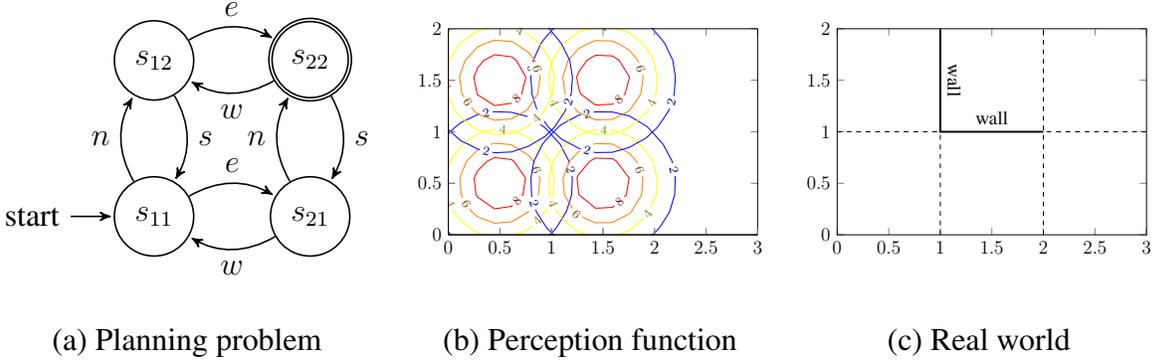
\begin{figure*}
  \begin{tabular}{ccc}
\begin{tikzpicture}[->,>=stealth',shorten >=1pt,auto,node distance=1cm,
                    semithick,  baseline=(current bounding box.center)]
                    \node[initial,state] (s11) {$s_{11}$};
\node[state,above = of s11] (s12) {$s_{12}$};
\node[state,right = of s11] (s21) {$s_{21}$};
\node[state,accepting,above = of s21] (s22) {$s_{22}$};
\path (s11) edge [bend left] node {$e$} (s21)
            edge [bend left] node {$n$} (s12)
      (s12) edge [bend left] node {$e$} (s22)
            edge [bend left] node {$s$} (s11)
      (s21) edge [bend left] node {$w$} (s11)
            edge [bend left] node {$n$} (s22)
      (s22) edge [bend left] node {$w$} (s12)
            edge [bend left] node {$s$} (s21);
          \end{tikzpicture} & 
\begin{tikzpicture}[scale=.6,baseline=(current bounding box.center)]
  \begin{axis}[view={0}{90},
    colormap/hot,
    xmin = 0,
    xmax = 3,
    ymin = 0,
    ymax = 2,
    axis equal image,
    samples=40,
    domain=0:2,
    y domain=0:2,
    ]
   \def\centerx{.5}
   \def\centery{.5}
    \addplot3[contour gnuplot,domain=-2:6,domain y=-5:3] 
        {10*exp(-10*( (x-\centerx)^2 + (y-\centery)^2)/3 )};
   \def\centerx{1.5}
   \def\centery{.5}
    \addplot3[contour gnuplot,domain=-2:6,domain y=-5:3] 
        {10*exp(-10*( (x-\centerx)^2 + (y-\centery)^2)/3 )};
   \def\centerx{1.5}
   \def\centery{1.5}
    \addplot3[contour gnuplot,domain=-2:6,domain y=-5:3] 
        {10*exp(-10*( (x-\centerx)^2 + (y-\centery)^2)/3 )};
   \def\centerx{.5}
   \def\centery{1.5}
    \addplot3[contour gnuplot,domain=-2:6,domain y=-5:3] 
        {10*exp(-10*( (x-\centerx)^2 + (y-\centery)^2)/3 )};
      \end{axis}
    \end{tikzpicture}
&       
\begin{tikzpicture}[scale=.6,baseline=(current bounding box.center)]
  \begin{axis}[view={0}{90},
    xmin = 0,
    xmax = 3,
    ymin = 0,
    ymax = 2,
    axis equal image,
    domain=0:2,
    y domain=0:2]
    \draw[very thick] (1,2) -- node[sloped,above]{wall} (1,1) -- node[auto]{wall}(2,1);
    \draw[thin,dashed] (0,1) -- (3,1);
    \draw[thin,dashed] (1,0) -- (1,2);
    \draw[thin,dashed] (2,0) -- (2,2);    
  \end{axis}
\end{tikzpicture} \\ \\ 
    (a) Planning problem & (b) Perception function & (c) Real world \\ 
  \end{tabular}
  \caption{\label{fig:pd_pf_w} (a) A planning problem on a domain
    composed of 4 states, corresponding to 4 rooms, no walls between
    them, and 4 actions $n$, $s$, $w$, and $e$ (go north, south, west,
    and east). Transitions that don't change the state are not shown.
    (b) A perception function associated to the planning domain.  (c)
    The real world: the building has 6 (and no 4) rooms, and two
    walls}
\end{figure*}

\section{Planning, Acting, and Learning}
%
A \emph{(deterministic) planning domain}
is a triple $\PD=\left<S,A,\gamma\right>$, composed of a
finite non empty set of states $S$, a finite non empty set of actions $A$, and
a state transition function $\gamma: S \times A \rightarrow S$.
A \emph{planning problem} is a triple
$\PP= \left<\PD,s_0,S_g\right>$ composed of a planning domain $\PD$, an
initial state $s_0 \in S$ and a set of goal states $S_g \subseteq S$.  
A plan $\pi$ for $\PD$ is a policy, i.e., a
partial function from $S$ to $A$.  
In discrete domains, each state $s \in S$ is represented 
with a set of (possibly multi-valued) \emph{state variables} ranging over a finite set of values.
A state $s \in S$ is a total assignment of values to the state variables.

The way in which an agent perceives the world is modeled by a
\emph{perception function}, i.e., a function
$f:\R^n\times S\rightarrow R^+$, defined as $f(\bx,s) = p(\bx|s)$,
where $p(\bx | s)=\frac{p(\bx,s)}{p(s)}$, and $p(\bx,s)$ is a joint
Probability Density Function (PDF) on $\R^n\times S$. 
In other words, $f(\bx,s)$ is the likelihood of observing $\bx$ being
in a state $s$. We call the $x_i$ of $\bx$ \emph{perception variables}. 

%
\begin{definition}[Extended planning domain]
  An \emph{extended planning domain} is a pair $\langle\PD,f\rangle$
  where $\PD$ is a planning domain and $f$ a perception function on the
  states of $\PD$. 
\end{definition}
Hereafter, if not explicitly specified, with ``planning domain'' we
will refer to extended planning domain.
\begin{example}
  \label{ex:simple}
  A simple planning domain with four states is shown in
  Figure~\ref{fig:pd_pf_w}. The transition system is shown in
  Figure~\ref{fig:pd_pf_w}.(a), the 
  relative perception function is shown in Figure~\ref{fig:pd_pf_w}.(b), and
  the real world, composed of a $3\times 2$ building is shown in Figure~\ref{fig:pd_pf_w}.(c).
  Each state $s_{ij}$ can be represented with two state variables
   $i,j$ taking values in $\{1,2\}$.
  The perception function $f(\langle x, y \rangle,s_{ij})$, shown in
  Figure~\ref{fig:pd_pf_w}.(b), is $f(\left<x,y\right>,s)=p(\left<x,y\right>\mid
s)$ where $p(\langle x,y\rangle|s_{ij}) = 
\mathcal{N}\left(\bmu=\left<i-0.5,j-0.5\right>,\bSigma=\left(\begin{smallmatrix} 1 & 0 \\
  0 & 1\end{smallmatrix}\right)\right)$.
Notice that, the agent's planning domain in Figure~\ref{fig:pd_pf_w}.(a-b)
is not coherent with the real world of 
Figure~\ref{fig:pd_pf_w}.(c), since  the 
transitions from $s_{12}$ and $s_{21}$ to $s_{22}$ are
not possible in the real world, due to the presence of walls.
In addition, there are no states corresponding to the rightmost part
of the building. This prevents the agent from reaching the goal.
Indeed, to reach the goal the agent should extend its planning domain,
as shown in Figure~\ref{fig:pd_pf_w_rev}, and plan for its actions in this
new planning domain. 
\end{example}


We now introduce an algorithm that interleaves planning,
acting and learning. Not only it is able to learn/update transitions
between existing states of the planning domain, but it can also
learn/update the perception function, and properly extend the planning
domain with new states. 
Algorithm \ref{PlanActLearn} \textsc{PlanActLearn} (PAL)  takes in input a
planning problem and a perception function. 
%
%
\begin{algorithm}[t]\footnotesize
\caption{\sc PlanActLearn - PAL}
\label{PlanActLearn}  
 \begin{algorithmic}[1] 
   \REQUIRE $\PP=\left<\PDdef,s_0,S_g,f\right>$ \COMMENT{A planning
     problem with a perception function}
   \REQUIRE{$p_{init}(\cdot)$ initialization for $f(\cdot,s)$}
   \STATE{$\Tr\gets\langle\rangle$}\ \ \ \ \ \COMMENT{The empty history of transitions}
   \STATE{$\Obs\gets\langle\rangle$}\ \ \ \ \ \COMMENT{The empty history of observations}
   \WHILE{$s_0\not\in S_g$}
   \STATE $\pi\gets plan(\PP)$
   \WHILE {$\pi(s_0)$ is defined and $\gamma$ has not been changed} 
   \STATE $\bx\gets act(\pi(s_0))$ 
   \STATE $s'_0 \gets \arg\!\max_{s\in S}f(\bx,s)$ \label{max-likelihood}
   \IF{$f(\bx,s_0') < (1-\epsilon)\cdot\max p_{init}(\cdot)$}
   
   \STATE $S\gets S\cup S_{new}$
   \STATE $s'_0\gets s_{new}\in S_{new}$
   \STATE Initialize $f(\cdot,s)$ for all $s\in S_{new}$
   \label{pinit}
   \ENDIF
   \STATE{$\Tr \gets append(\Tr,\left<s_0,\pi(s_0),s_0'\right>)$}
   \COMMENT{extend the transition history with the last one}
   \STATE{$\Obs \gets append(\Obs,\left<s'_0,\bx\right>$)}
   \COMMENT{extend the observation history with the last one}
   \STATE $\gamma \gets \updgamma(\gamma,\Tr)$
   \STATE $f \gets \updf(f,\Obs)$
   \STATE $ s_0 \gets s'_0$
   \ENDWHILE   
   \ENDWHILE
\end{algorithmic}
\end{algorithm}
%
%
At line 4, $plan(\PP)$ generates a plan $\pi$ by applying some planning algorithm for deterministic domains\footnote{We assume that the sequential plan returned by the planning algorithm can be transformed into a policy $\pi$. 
Since here we plan for reachability goals, sequences of actions can be mapped into policies}.
If $plan(\PP)$ does not find a plan to the goal, then it generates a
plan to learn the domain, e.g., a policy that explores
  unknown regions of the planning domain.
We then execute the planned action $\pi(s_0)$ in the current state
$s_0$, and  observe the world through the perception
  variables in $\bx$ (line 6).
We then determine the state $s'_0$ that maximizes the likelihood of
observing $\bx$ (line 7).

Explicitly computing $\arg\!\max_{s\in S}f(\bx,s)$ requires
  to compute the perception function for each state $s\in S$. This
  might be prohibitive. An approximated solution can be obtained by
  using a greedy best first search algorithm over the set of states,
  starting from $\gamma(s_0,\pi(s_0))$. We adopt this approximation in
  our experimental evaluation.

  If the maximum likelihood is below the threshold
  \mbox{$(1-\epsilon)\cdot\max p_{init}(\cdot)$}, with
  $\epsilon \in [0,1]$, then we extend the set of states $S$ with a
  new set of states $S_{new}$, and we select a new state $s_{new}$
  from it (lines 9--10). We then initialise its perception function
  $f(\cdot,s_{new})$ with $p_{init}(\cdot)$ (line 11).  Notice that,
  low values of $\epsilon$, promote the easy introduction of new
  states, while with high values of $\epsilon$ we are cautious in
  creating new states.

If states are represented with more than one state variable, $s_{new}$
can be generated, either by extending the set of state variables, or
by extending the values of existing state variables.  Adding a new
variable, will result in multiplying the size of $S$ by the
cardinality of the domain of the new variable. E.g., adding a boolean
variable will make $|S_{new}|=|S|$, and it will double the number of
states. Instead, extending the set of values of one variable $v$ with
a new value will result in producing less states. Indeed, in this
case, the number of states are multiplied by
$\frac{|\dom_i|+1}{|\dom_i|}$, were $\dom_i$ is the set of values of
the $i$-th state variable.  However, it is not obvious which variable
should be extended.  A simple heuristic could be to extend the
variable that is maximally affected by $\pi(s_0)$ according to
$\gamma$. More formally, we select the $i^*$-th variable where
\begin{equation}
  \label{eq:best variable to extend}
i^* =\argmax_{i\in\{1,\dots,m\}}\left(\sum_{s\in S}\mathbb{1}_{s[i] \neq \gamma(\pi(s_0),a)[i]}\right)
\end{equation}
where $m$ is the number of state variables, $s[i]$ is the value of the
$i$-th state variable in $s$, and
$\mathbb{1}_{cond}$ is equal to $1$ if \emph{cond} is true, $0$
otherwise.
By extending the domain of the $i^*$-variable with a new value $v_{new}$, we generate a new set
of states
\begin{equation}
  \label{eq:S_new}
  S_{new}=\{s[i^*:=v_{new}] \mid s\in S\}
\end{equation}
We select $s_{new}=s_0[i^*:=v_{new}]$.
\footnote{When states are represented with state
  variables, $s[i:=v]$ denotes the state obtained by assigning the
  value $v$ to the $i$-th variable of $s$ and leaving all the other
  variables unchanged.}
For instance, in the example, if we have executed action $e$ in
$s_{21}$, we extend the first variable, since 
$e$ changes only the value of the first state variable,
and generate the set of new states $S_{new}=\{s_{31},s_{32}\},$

We then extend the sequence of transitions $\Tr$ and of observations $\Obs$, and learn the  new transition function $\gamma$ and the new perception function $f$. 
The functions $\updgamma$ and $\updf$ update the transition function
$\gamma$ and the perception function $f$, respectively, depending on
the data available in $\Tr$ and $\Obs$. The update functions take into
account (i) the current model, (ii)
what has been observed in the past, i.e., $\Tr$ and $\Obs$, and (iii)
what has been just observed, i.e., $\left<s_0,\pi(s_0),s_0'\right>$
and $\left<s_0',\bx\right>$.
%
The update functions can be defined in several different ways, depending
on whether we follow a cautious strategy, where changes are made only
if there is a certain number of evidences from acting and perceiving
the real world, or a more impulsive reaction to what the agent has
just observed.
%


\noindent
\textbf{Updating transitions:}
$\updgamma$ decides whether and how to update the transition function.
Suppose that, after executing the action $a$ from the state $s_0$,
the agent perceives $\bx$, and suppose that
$s'_0=\argmax_{s}(f(\bx,s))$,
i.e., the most likely reached state, 
is different from the state predicted by the agent planning domain,
i.e., $s'_0\neq \gamma(a,s_0)$, then $\gamma$ may need to be revised to take into account this discrepancy. 
Since our domain is deterministic (the transition $\gamma$ must lead
to a single state), if the execution of an action leads to an
unexpected state, we have only two options: either change $\gamma$
with the new transition or not. 
We propose the following transition update function that
depends on $\alpha$: 
We define 
$\updgamma(\gamma,\Tr)(s,a)=s'$  where $s'$ is \emph{one element}%
of the set
\begin{equation}
\{\argmax_{s'\in S}\left(\alpha\cdot\mathbb{1}_{s'=\gamma(s,a)}+
    (1-\alpha)\cdot|\{i\mid \Tr_i=\left<s,a,s'\right>\}|
\right)\}
\label{eq:upd_gamma}
\end{equation}
where $\Tr_i$ is the $i$-th element of
$\Tr$, and $\alpha\in[0,1]$. 
Notice that, if
$\alpha = 1$, we are extremely cautious, we strongly believe in our
model of the world, and we never change the transition $\gamma$.
Conversely, if $\alpha = 0$, we are extremely impulsive, we do not trust our model,
and just one evidence makes us to change the model. In the
intermediate cases, $\alpha \in (0,1)$, depending on the value of
$\alpha$, we need more or less evidence to change the planning
domain.

\noindent
\textbf{Updating the perception function:}  
The update of the perception function is based on the current
perception function $f(\bx,s)$ for $s\in S$ and the set of
observations $\Obs$. We suppose that the perception function is parametric on
$\prm=\left<\theta_1,\dots,\theta_k\right>$. In Example~\ref{ex:simple}, $\prm=\langle\theta_1,\theta_2\rangle$ with $\theta_1=\bmu$ and $\theta_2=\bSigma$, i.e., the mean and the covariance
matrix of the normal distribution associated to any state.
Given a new observation $\left<\bx,s\right>$ and a set of previous
observations $\Obs(s)=\left<\bx^{(0)},\dots, \bx^{(k)}\right>$ about an
abstract state $s\in S$, we have to update the parameters $\prm_s$ of
the perception function $f(\cdot,s)$ in order to
maximize the likelihood of the entire set of observations extended
with the new observation. Also in this case the agent can be more or less
careful in the revision. This is expressed by a parameter
$\beta\in[0,1]$, where, the higher the value of $\beta$ the
more careful the agent is in
the revision. If $f(\bx,s) = p(\bx|s,\prm_s)$, we define
$\updf(f,\Obs)(\bx,s)$ as 
$p(\bx\mid s,\prm'_s))$
where: 
\begin{align}\small
  \label{eq:update-f}
  \prm'_s & = \beta\cdot\prm_s+(1-\beta)\cdot
            \argmax_{\prm''} \mathcal{L}(\prm_s'',\Obs(s),\bx,s)
\end{align}
where 
$\mathcal{L}(\prm,\bx^{(1)},\dots,\bx^{(n)},s)$ 
is the likelihood of the 
parameters $\prm$ for the observations $\bx^{(1)},\dots,\bx^{(n)}$,
defined as: 
\begin{align}
  \label{eq:max-like-prm}
\mathcal{L}(\prm,\bx^{(1)},\dots,\bx^{(n)},s) & = 
  \prod_{i=1}^nP(\bx^{(i)}|s,\prm)
\end{align}
Intuitively Equation \eqref{eq:update-f} defines the parameters
$\prm'_s$ of the updated perception function for a state $s$ as a
convex combination, based on the parameter $\beta$, of the parameters
of the previous perception function for $s$, i.e., $\prm_s$ and the
parameters $\prm''$ that maximize the likelihood of the past and
current observations about state $s$ (equation
\eqref{eq:max-like-prm}).  An efficient procedure for incremental
estimation of the second term of \eqref{eq:update-f}, is described in
\cite{bishop2006}. In case of Multivariate Gaussian distribution,
$\prm_s$ contains the mean $\bmu_s$ and covariance matrix $\bSigma_s$,
and the updates defined in equation \eqref{eq:update-f} can be
efficiently computed as follows:
\begin{align*}
  \bmu'_s & = \beta\cdot\bmu_s+(1-\beta)(\bmu_s + \Delta\bmu_s) \\
  \bSigma'_s & = \beta\cdot\bSigma_s+(1-\beta)(\bSigma_s + \Delta\bSigma_s) \\
\end{align*}
where $\Delta\bmu_s=\frac{1}{|\Obs(s)|}(\bx-\bmu_s)$
and $\Delta\bSigma^2_s = \frac{1}{|O(s)|}(\bx-\bmu_s')^2 +
\frac{|O(s)|-1}{|O(s)|}(\Delta\bmu_s^2-2\bmu_s\Delta\bmu_s) -\frac{1}{|O(s)|}\bSigma^2_s$.
Concerning the parameter $\beta\in[0,1]$, it plays the similar role
as $\alpha$ in the case of the revision of the transition function.
It balances the update depending on whether the agent is cautious or impulsive
about the current perception function, and the new perceptions. 

  Notice that \PAL\ could implement specific planning strategies to
  learn the perception function for newly introduced states and states
  with few observations in $\Obs$. This allows the agent to learn new
  perception functions from experience.


\begin{example} 
Let us now describe how our algorithm works in Example~\ref{ex:simple}
and how the goal is reached by creating new states and changing the
model to the one described in Figure~\ref{fig:pd_pf_w_rev}.
\begin{enumerate}
\item 
  Suppose that the robot is initially in the position $(0.5,0.5)$ and
  that, according to its perception function, it believes to be in
  $s_{11}$ (notice indeed that $s_{11}=\argmax_{s_{ij}}f((0.5,0.5),s_{ij})$
  when
  $f((x,y),s_{ij})=\N((0.5,0.5),\bmu=(i-0.5,j-0.5),
  \bSigma=\left(\begin{smallmatrix}1 & 0 \\ 0 & 1\end{smallmatrix}\right))$). 
\item (line 4)
According to the planning domain in Figure~\ref{fig:pd_pf_w}.(a), $Plan(\PP)$ 
can generate two plans, the one that reaches the goal passing through
$s_{21}$ and the one that passes through $s_{12}$. Let us suppose
that it generates the former, i.e., the plan $\pi(s_{11})=e$ and
$\pi(s_{21})=n$.

\item (line 6-7)
Since $\pi(s_{11})$ is defined, we execute the action $e$, which moves
the robot of one unit in the east direction, and returns the current
position in $\bx$, which will be some value close to
$\left<1.5,0.5\right>$. Notice that we cannot assume that $\bx$ is
exactly $\left<1.5,0.5\right>$, since we have to take into
consideration that sensors and actuators can be noisy. So suppose that
the observed values after the execution of $\pi(s_{11})$ are 
$\left<1.51,0.49\right>$. 
Given the current $f$, the state $s$ that maximizes 
$f(\bx,s)$ is $s_{21}$, therefore $s_0' = s_{21}$.
\item (lines 8,13,14)
  Suppose that the condition on line 8 is false. We then do not create a new state. 
We add the transition to the history and we have $\Tr = \langle \langle s_{11}, e, s_{21} \rangle \rangle$.
Similarly we have $\Obs = \langle \langle s_{21}, \langle 1.51,0.49\rangle \rangle$. 
\item (lines 15) We then update the transition function: $\updgamma$
  does not produce any change, since $s_{21} =
  \gamma(s_{11},e)$. Indeed in this case the transition function
  $\gamma$ correctly predicts, at the abstract level, the effects of
  the execution of action $e$ in state $s_{11}$.

\item (line 16) The update of the perception function will
slightly move the mean $\bmu$, from $\left<1.5,0.5\right>$ in the
  direction of the current perception i.e., $\left<1.51,0.49\right>$
  and the $\bSigma$ will also be updated. 
\item \label{item_hit_the_wall} We then update $s_0'$ to $s_{21}$ and go back to (lines 3,4).
Since $\pi(s_{21})=n$, 
we execute the action moving one unit north from $s_{21}$. 
But the execution of this action does not have the effect that is expected by the agent, i.e., 
it does not reach state $s_{22}$. 
Indeed, the execution of $n$ starting from the position
$\left<1.51,0.49\right>$ would result in hitting the wall, 
the presence of which was not expected by the agent.
Let us suppose that the execution of this action will result in the robot
  doing nothing, and $act(\pi(s_{21}))$ will return the value $\bx$
  which is the same as the previous one i.e.,
  $\bx=\left<1.51,0.49\right>$. 
\item $s_{21}$ is the state that maximizes the observed
  $\bx$, and we proceed as before, by not generating a new state and
  appending the new transition to $\Tr$ such that
  $\Tr = \langle 
  \langle s_{11}, e, s_{21} \rangle,
  \langle s_{21}, n,  s_{21} \rangle \rangle$
  while $\Obs$ becomes
  $\langle
  \langle s_{21}, \langle 1.51,0.49\rangle\rangle,
  \langle s_{21}, \langle 1.51,0.49\rangle\rangle
  \rangle
  $
\item (line 15) The transition function this time gets updated in different
  ways depending on the value of $\alpha$.
  Let's compute the arguments of the argmax of equation 
  \eqref{eq:upd_gamma} with $a=n$ and $s=s_{21}$;
  $$\small 
  \begin{array}{|l|l|}\hline
    s' & \alpha\cdot\mathbb{1}_{s'=\gamma(s_{21},n)}+
         (1-\alpha)\cdot|\{i\mid \Tr_i=\left<s_{21},n,s'\right>|\\
    \hline 
    s_{11} & \alpha\cdot 0 + (1-\alpha)\cdot 0 = 0 \\
    s_{21} & \alpha\cdot 0 + (1-\alpha)\cdot 1 = (1-\alpha)\\
    s_{12} & \alpha\cdot 0 + (1-\alpha)\cdot 0 = \alpha\\
    s_{22} & \alpha\cdot 1 + (1-\alpha)\cdot 0 = 0 \\ \hline              
  \end{array}
  $$
If $\alpha < 1/2$, 
we are reasonably keen to learn from acting in the
real world that the state that maximizes equation \eqref{eq:upd_gamma} is $s_{21}$
and $\updgamma$ deletes $\gamma(s_{21},n) = s_{22}$ and adds
$\gamma(s_{21},n) = s_{21}$, i.e., the agent understands that there is
a wall that does not allow the robot to move north from state
$s_{21}$.
If instead $\alpha > 1/2$, then the state that maximizes \eqref{eq:upd_gamma}
is $s_{22}$ and $\gamma(s_{21},n)=s_{22}$ will be
kept. Notice that after $k$ attempts to execute the actions $n$ in
state $s_{21}$ without updating the transition function, in order to
change the transition function it is enough to have $\alpha <
\frac{k}{1+k}$. So if $\alpha\neq 1$, sooner or later the agent will
update $\gamma$. 
\item At this point we go back to $Plan(\PP)$ which generates the
  alternative plan that passes through $s_{12}$, and sends the robot
  back to state $s_{11}$ and then to state $s_{12}$, in a similar way
  to what happened in the case of going through $s_{21}$.
\item At this point the planning domain of the agent is shown
  below. Notice that the goal is not reachable. 
\begin{center}
  \begin{tikzpicture}[->,>=stealth',shorten >=1pt,auto,node distance=.6cm,
                    semithick,  baseline=(current bounding box.center)]
                    \node[initial,state] (s11) {$s_{11}$};
\node[state,above = of s11] (s12) {$s_{12}$};
\node[state,right = of s11] (s21) {$s_{21}$};
\node[state,accepting,above = of s21] (s22) {$s_{22}$};
\path (s11) edge [bend left] node {$e$} (s21)
            edge [bend left] node {$n$} (s12)
      (s12) edge [bend left] node {$s$} (s11)
      (s21) edge [bend left] node {$w$} (s11)
      (s22) edge node {$w$} (s12)
            edge node {$s$} (s21);            
     \end{tikzpicture} 
\end{center}
        
\item After having explored all the possibilities without reaching the
  goal, $Plan(\PP)$ generates an exploration plan. Suppose that it
  generates a plan $\pi$ with $\pi(s_{21})=e$. The observation after
  the execution of such a $\pi$ returns $\bx = \langle x,y \rangle$
  close to $\left<2.5,0.5\right>$.  In that position $s_{21}$
  maximises $f(\bx,s)$, however, suppose that such a value is below
  the threshold $(1-\epsilon)\cdot \max p_{init}(\cdot)$. We therefore
  have to extend the set of states. If states are represented with two 
  variables, according to equation \eqref{eq:best variable to extend},
  we will extend the first variable (the one modified by
  $e$) introducing a new value, resulting in
  $S_{new}=\{s_{31},s_{32}\}$ (see equation \eqref{eq:S_new}) and we
  set $s_{new}$ to $s_{31}$. 
  If instead, we introduce a new (third)
  boolean variable we get a domain with eight states. 
\item 
$\Tr$ gets updated by adding the transition 
$\langle s_{21},e,s_{31} \rangle$, and $\Obs$ by adding 
the pair $\langle s_{31}, \bx \rangle$.
The update function $\updgamma$ may create the new transition
$\gamma(s_{21},e) = s_{31}$ (if $\alpha$ is small enough) and $\updf$ 
will initialize the perception function $f(\cdot,s_{31})$ with
$p_{init}\sim\N(\bmu,\bSigma)$ with $\bmu=\bx$, and
$\bSigma=\left(\begin{smallmatrix}.1&0\\0&.1\end{smallmatrix}\right)$. 

\item In the next step, since there is no plan from the newly added
  state $s_{31}$, $plan(\PP)$ tries to explore the domain and the
  transition $\gamma(s_{31},n)$ and the corresponding $f$ are
  learned. Since no plan to the goal exists yet, while trying to learn
  the domain, $plan(\PP)$ may add the new transitions
  $\gamma(s_{31},w)=s_{21}$ and $\gamma(s_{32},s)=s_{31}$
\item
In the final step, $plan(\PP)$ learns the transition
$\gamma(s_{32},w)=s_{22}$, and finally finds the plan to the goal
$\pi(s_{32}) = w$. Furthermore, the agent has
updated its initial planning domain, obtaining the planning domain
shown in Figure~\ref{fig:pd_pf_w_rev}.
Notice that this planning
domain is not completely correct, as there are no information about the execution of actions in $s_{22}$. This is due to the fact that, in this simple example, the agent has planned no
actions in $s_{22}$ (since it was the goal) and therefore it has not
learned anything about the transitions and the perceptions functions
of this node. 
\end{enumerate}
\begin{figure}
  \begin{center}
\begin{tikzpicture}[->,>=stealth',shorten >=1pt,auto,node distance=.6cm,
                    semithick,  baseline=(current bounding box.center)]
\node[initial,state] (s11) {$s_{11}$};
\node[state,above = of s11] (s12) {$s_{12}$};
\node[state,right = of s11] (s21) {$s_{21}$};
\node[state,accepting,above = of s21] (s22) {$s_{22}$};
\node[state,right = of s21] (s31) {$s_{31}$};
\node[state,right = of s22] (s32) {$s_{32}$};
\path (s11) edge [bend left] node {$e$} (s21)
            edge [bend left] node {$n$} (s12)
      (s12) edge [bend left] node {$s$} (s11)
      (s21) edge [bend left] node {$e$} (s31)
            edge [bend left] node {$w$} (s11)
      (s22) edge node {$w$} (s12)
            edge node {$s$} (s21)
      (s31) edge [bend left] node {$w$} (s21)
            edge [bend left] node {$n$} (s32)
      (s32) edge node {$w$} (s22)
            edge [bend left] node {$s$} (s31);
experiments
      \end{tikzpicture} 
\end{center}      
  \caption{\label{fig:pd_pf_w_rev} The new planning domain
    obtained by extending the initial domain of Figure~\ref{fig:pd_pf_w},
    with two new states}
\vspace*{-10pt}
\end{figure}
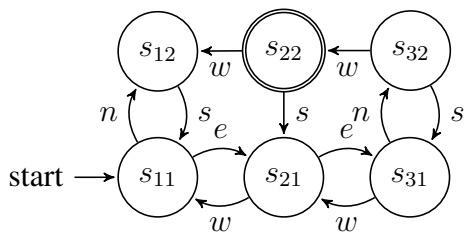
\end{example}

A remark is in order.  \PAL\ stops when the agent
  perceives to be in a goal state. However this perception might be
  erroneous. For instance the agent might perceive to be in $s_{22}$
  (the goal) even if it is in a different room close to the goal.
  Even if the current \PAL\ algorithm stops when the agent perceives
  to be in a goal state, our framework opens up the possibility for
  the agent to continue executing further actions to confirm its
  perception.  

 In summary, our approach introduces three parameters,
  each of them representing how the agent trust the three key
  components of its model: $\epsilon$ for states, $\alpha$ for
  transitions, and $\beta$ for perception functions.  They are
  conceptually independent. However, there are reasonable criteria to
  discard certain parameter combinations. For instance, low
  $\epsilon$, and $\alpha$ and $\beta$ close to 1 will generate many
  close and unconnected states. See the Section ``Experimental
  Evaluation" for a thorough analysis of how setting of these
  parameters influence the whole process.

\section{Measuring the coherence of the model}
In order to estimate the quality of the model generated by the \PAL\
algorithm, we should define a method to measure the coherence between
an abstract model with perception function and the real world.

We introduce a measure called \emph{divergence}. 
Intuitively, a low divergence means that if
$\gamma(a,s) = s'$, then if the agent perceives to be in $s$ and
performs $a$, then after the execution of $a$ it will perceive to
be in the state $s'$.

We suppose to have a stochastic model of the real execution of actions. 
Under Markovian hypothesis, every action $a \in A$ can be
modeled as a conditional PDF $p_a(\bx'|\bx)$,
which expresses the probability of measuring $\bx'$ after executing the action $a$ in a
state in which the agent perceives $\bx$. 
It represents
the effects of executing the action $a$ in the real world.

To measure the quality of the abstract planning domain, 
we have to compare $p_a$ with how
the action $a$ is modeled in the domain. 
Suppose that an agent perceives $\bx$, and that the state $s$
maximizes the likelihood of perceiving $\bx$. 
Suppose that the action $a$ is executed.
According to its abstract model, the agent will believe to be in
the state $s'=\gamma(a,s)$.
After the actual execution of action $a$, it will perceive $\bx'$
with a probability $p_a(\bx'|\bx)$. 
However, according to the agent's abstract model, 
the probability of observing $\bx'$ after the execution
of $a$ is $p(\bx'|s')$. 
The closer the two distributions are, 
the more coherent the abstract representation is. 
To estimate how well $p(\bx'|s')$ approximates the real distribution $p_a(\bx'|\bx)$, we
use the notion of  {\em divergence}, which is the opposite notion of coherence
(the lower the divergence, the higher the coherence), 
and we formalize it  with the 
{\em KL divergence} 
$KL(p_a(\bx'|\bx)||p(\bx'|s'))$,
defined as: 
$$
\int_{\bx'}p_a(\bx'|\bx)\log\left(\frac{p_a(\bx'|\bx)}{p(\bx'|s')}\right)\:d\bx'
$$
We can therefore define the divergence measure as
\begin{align}
  \label{eq:coherence}
  \int_{\bx}\sum_{a\in
  A}\mathrm{KL}(p_a(\bx'|\bx)||p(\bx'|\gamma(a,s_{\bx})))\cdot p_A(\bx)\: d\bx
\end{align}
where $s_{\bx}=\argmax_{s\in  S}f(\bx,s))$ and $p_A(\bx)$ is a distribution of all the possible perceptions
that can be obtained by the agent following all the possible sequences
of actions, i.e., 
$$
p_A(\bx)=\sum_{\left<a_1.\dots,a_n\right>\in A^+}p_{a_n}(\bx|\bx^{(n-1)})\cdot\prod_{i=1}^{n-1}p_{a_i}(\bx^{(i)}|\bx^{(i-1)})
$$
\def\ba{\mathbf{a}}
where $A^+$ is the set of finite non empty sequences of actions in $A$
and $\bx^{(0)}$ is the perception of the agent in the initial state. 
However, computing \eqref{eq:coherence} analytically is very difficult.
We therefore
estimate \eqref{eq:coherence} by random walk sampling method. Starting
from an initial observation $\bx^{(0)}$ we generate $N$ random
walks $\ba_1,\dots,\ba_N$, with
$\ba_i=\langle a_{i,1},\dots,a_{i,n_i}\rangle$ and sample $\bx^{(i)}$
from $\prod_{j=1}^{n_i}p_{a_{ij}}(\bx_{j}|\bx_{j-1})$. 
We approximate \eqref{eq:coherence} with
\begin{align}
  \label{eq:coherence-approx}
\frac{1}{N}\sum_{k=1}^N\sum_{a\in A}\mathrm{KL}(p_a(\bx'|\bx^{(k)})|| p(\bx'|\gamma(a,s^{(k)})))
\end{align}
where $s^{(k)} = \argmax_{s\in S}f(\bx^{(k)},s)$ for $1\leq k \leq
n$. 
In our specific example, since we are working with Gaussian
distributions, we have that 
$p_a(\bx'|\bx) =
\N(\bmu=a(\bx),\bSigma=\bSigma_a)$,
where $a(\bx)$ 
is some real function that maps  
$\bx$ in the expected value $a(\bx)$ after performing the action $a$, and $\bSigma_a$ is the model of the noise of the
sensors/actuators associated to $a$. For instance, in Example~\ref{ex:simple}
$$
e(\langle x,y\rangle)=
\begin{cases}
  \langle x+1,y\rangle & \parbox[t]{.3\textwidth}{If there are no walls
    between \\ $\left<x,y\right>$ ad $\left<x+1,y\right>$} \\
    \langle x,y\rangle & \mbox{Otherwise}
  \end{cases}
  $$
Furthermore, the KL divergence of Multivariate Gaussians can be
computed analytically. 


\section{Experimental Evaluation}

We propose three experiments. We first run \PAL\ on Example
\ref{ex:simple} to give a first intuitive idea on how the parameters
$\alpha$,$\beta$, and $\epsilon$ influence the learning of the model.  Successively we run \PAL\ on a
larger artificial test case, to give a first experimental evidence how
\PAL\ converges to a coherent model.  In the final experiment, we
verify the scalability of the approach. We let \PAL\ create a modelwith a large number
of states (more than 3,000,000) and we measure the time necessary to
run each plan-act-learn loop, including the time to compute the state
with maximum likelihood. 
We implement our approach using a planner based on A$^*$ algorithm,
with heuristic based on the Euclidean distance from the goal, and we
compute an estimation of the maximum likelihood, by using a greedy
best first algorithm. 

The prototype implementation uses a simple heuristic for the
exploration phase, which avoids to repeat the same action and reaching
recently visited states. We run the first two experiments with a
single state variable, and the third one with three state variables.

In the first experiment, the initial planning
domain is shown in Figure~\ref{fig:pd_pf_w} and, for different
configurations of the parameters $\alpha$, $\beta$, and $\epsilon$ in
$\{0.0,0.5, 1.0\}$, we run the \PAL\ algorithm 10 times. We measure
the average number of states of the final model ($|S|$), the
reduction/improvement of the divergence (``\% lrn") and the percentage of
achieved goals (\%G). The results are reported in table
~\ref{fig:explanatory-exp-eval}

Consider first the effects of the parameter $\epsilon$:
\begin{itemize}
\item $\epsilon = 1$ prevents the creation of new states. Indeed, in all cases, 
no new state is created, and, as expected, the learned model is not more coherent than the initial one - the percentage of learning ``\% lrn'' ranges from a negative number ($-11.7$) to very low improvement ($0.26$).  Indeed, without
  creating new states, \PAL\ never understands that there are new
  rooms.  Because of this lack of coherence, in many cases \PAL\ does
  not manage to reach the goal within the given timeout (100 steps).
  The reason why in some cases it manages to reach the goal is simply
  due to the fact that, when no plan exists according to the model,
  then a random policy is tried, which in some cases reaches the goal
  by chance, due to the simple and small domain.
\item $\epsilon = 0$  tends to create many new states:
  $|S| \in [20.2,32.9]$.  In spite of this, when $\alpha = 0$, the
  learning is much better than when no new states are created (``\%
  lrn" $\in [0.49,0.54]$) and the goal is often reached. The learning
  gets worse by increasing $\alpha$, since we learn many new states
  that are however scarcely connected to the states in the initial
  model.
\item $\epsilon = 0.5$ represents a balanced situation.  The number of
  new learned states is the right one ($|S| \approx 6$) for all the
  values of the other parameters.  Moreover, with $\alpha = 0$ we have
  the best learning of a coherent model (``\% lrn $= 0.72$)), since
  we allow the update of the transition function by connecting the two
  new states with the four initial ones.  The performance of learning
  smoothly decreases by increasing $\alpha$ to $0.5$, while it becomes
  low in the case of $\alpha = 1$, due to the fact that the new states are not connected with the old ones. 
\end{itemize}
In the case $\alpha = 0$ and $\alpha = 0.5$, the parameter $\beta$,
when it is low ($\beta = 0$), tends to decrease the amount of learning
towards a coherent model, by producing the two worst results (``\%
lrn" $=-11.17$ and $-4.71$) in the case $\epsilon = 1$. This is
because, since we cannot learn new states, with a low $\beta$ we allow
the perception function to move the same old states to different
positions, thus creating a rather incoherent model.
\begin{table}[t]\footnotesize
\centering
\def\toprule{\hline}
\def\midrule{\hline\hline}
\def\bottomrule{\hline}
\tabcolsep=3pt
\begin{tabular}{|l|l|r|r|r|r|r|r|r|r|r|}
\toprule
$\alpha$ & $\beta$ & 
\multicolumn{3}{|c|}{$\epsilon=0.0$} & 
\multicolumn{3}{|c|}{$\epsilon=0.5$} & 
\multicolumn{3}{|c|}{$\epsilon=1.0$} \\  \cline{3-11}
& & $|S|$ & \%lrn & \%G & 
$|S|$ & \%lrn & \%G & 
$|S|$ & \%lrn & \%G \\ \midrule
0.0 & 0.0 &          21.6 &        0.49 &           100 & 
                      6.0 &        0.72 &            70 & 
                      4.0 &       -4.71 &            10 \\
    & 0.5 &          24.7 &        0.50 &            90 & 
                      5.9 &        0.72 &            80 & 
                      4.0 &       -0.41 &            90 \\
    & 1.0 &          20.2 &        0.54 &            90 & 
                      5.9 &        0.72 &            70 & 
                      4.0 &        0.18 &            90 \\ 
0.5 & 0.0 &          25.8 &        0.19 &           100 & 
                      5.9 &        0.66 &            80 & 
                      4.0 &      -11.17 &            20 \\
    & 0.5 &          30.7 &        0.15 &            70 & 
                      5.9 &        0.69 &            80 & 
                      4.0 &        0.26 &            80 \\
    & 1.0 &          32.9 &        0.16 &            70 & 
                      6.0 &        0.63 &            80 & 
                      4.0 &        0.07 &            70 \\
1.0 & 0.0 &          25.0 &        0.15 &            90 & 
                      5.9 &        0.25 &            50 & 
                      4.0 &        0.01 &            10 \\
    & 0.5 &          24.4 &        0.18 &            80 & 
                      5.8 &        0.24 &            80 & 
                      4.0 &       -1.26 &            80 \\
    & 1.0 &          28.5 &        0.16 &            80 & 
                     6.0 &        0.27 &            100 & 
                      4.0 &        0.00 &            90 \\
\bottomrule
\end{tabular}
\caption{Performance of \PAL\ on Example~\ref{ex:simple} depending on
  $\alpha$, $\beta$, and $\epsilon$. Results are averaged on the 10
  runs. 
\label{fig:explanatory-exp-eval}}
\vspace*{-12pt}
\end{table}

In the second set of experiments, we consider a $5\times5$ building with
randomly generated walls completely unknown by the agent. Differently from the previous
experiments, we test the capability of \PAL\ to create a planning
domain from scratch, while it is trying to achieve 10 randomly
generated goals.
We initialise the agent with a model
containing only two states, i.e., $S=\{s_0,s_{g_0}\}$.  The mean
$\bmu_{s_0}$ of the initial state is set to $\left<0.5,0.5\right>$,
the mean of $s_{g_0}$ of the perception function of $s_{g_0}$ is
randomly generated.  The covariance matrixes $\bSigma_{s_0}$ and
$\bSigma_{s_{g_0}}$ are initialized to
$\left(\begin{smallmatrix}.1&0\\0&.1\end{smallmatrix}\right)$.  The
objective of the agent is to reach the goal $s_{g_0}$, and
successively to reach other 9 goals $s_{g_1},\dots,s_{g_9}$, which are
also randomly generated. We run this experiment, for every combination
of $\alpha$, $\beta$, and $\epsilon$ in $\{0.0,\ 0.5,\ 1.0\}$.  In
Figure \ref{fig:big_experiments} we report the divergence (in the
three plots on the left of the figure) and the number of states that
are generated (in the three plots on the right) depending on the time
used by \PAL\ to plan, act and learn (the x axis), and depending on
the parameters $\alpha$, $\beta$, and $\epsilon$.  Notice that the
graphs have different scales, since with a uniform scale some of the
graphs would not be readable.

If $\epsilon = 1$, \PAL\ cannot add new states to the planning model,
  and therefore, planning is useless, and the agent adopts a random
  walk strategy. Furthermore, the divergence is computed only on a single
  state. The consequence is that $\alpha$ does not have any effect,
  since with a single state there is no transition to revise. 
  Instead, the value of $\beta$ has the effect of (dis)allowing the change of
  the perception function associated to the single state $s_0$. If
  $\beta=1$, the perception function  $f(\bx,s_0)$ is not changed and,
  consequently, the divergence is constant (i.e., it takes its initial 
  value $\approx 5000$); with $\beta\neq 1$, instead, the
  perception function $f(\bx,s_0)$ is updated to take into 
  account the observations that the agent accumulates
  during its random walk, but after short time it converges to a
  constant value $\approx 13.0$. 

  If $\epsilon = 0$, \PAL\ tends to generate an eccessive number
  of states independently from $\alpha$ and $\beta$: We get to about $600$
  states in $800$ seconds. In this case \PAL\ learns a domain by
  decreasing significantly the divergence, which gets below $500$ for
  all the values of $\alpha$ and $\beta$. It takes however a long
  time to complete the tasks, up to 800 seconds, because the model is
  uselessly accurate. 
  
  If $\epsilon = 0.5$, \PAL\ generates a reasonable number of
  states.  For all the values of $\alpha$ and
  $\beta$, the completion of the task requires much less time than the
  case of $\epsilon = 0$.
  The best model, i.e., the one closest to our intuition, is the one
  generated in the case of $\epsilon = 0.5$, $\alpha = 0.5$, and
  $\beta = 0$. It has indeed 25 states, each one corresponding to the
  25 rooms in the building, and with transitions taking into account
  the walls.  In this case, the divergence rapidly decreases to values
  below 100.  Moreover, in all cases in which $\epsilon = 0.5$, we
  have divergences much lower than in the case $\epsilon = 0$ (please
  notice the different scale in the two graphs). Finally, we have
  lower divergences for low values of $\alpha$ ($\alpha = 0$ and
  $0.5$) than in the case of $\alpha = 1$, since, as usual,
  $\alpha = 1$ does not allow \PAL\ to connect the new states to the
  old ones.
  
  In the last experiment we test the scalability of the
  proposed approach. We run \PAL\ on a more complex domain. The goal
  is to move packs in a target position of a building. The building size
  is $10\times 10$ rooms, with 5 packs. We represent states with three
  state variables: two of them encode the rooms, and the third one
  encodes the number of packs the agent is carring.
  With proper parameters ($\alpha,\beta,\epsilon=0.5$) \PAL\ creates a
  model with about 60,0000 states, corresponding to the intuitive
  model. However, to further check the scalability, we set the parameters $\epsilon$
  close to 0 so that \PAL\ introduces a lot of extra states (about $3\cdot10^6$). For each
  size of $S$, we log the average time \PAL\ takes to execute
  one iteration of the plan-act-learn loop (lines 4--17 of Algorithm \ref{PlanActLearn}).
  The experiments log reveals a quasi linear dependency between the size
  of $S$ and the time necessary to run one iteration of the
  plan-act-learn loop. Notice that the last loop, which process about
  3 milion states, takes about 3 minutes. The total time for creating
  a model with 3 milion states is about 4.5 hours. 
\begin{figure}
\centering
\includegraphics[width=\columnwidth]{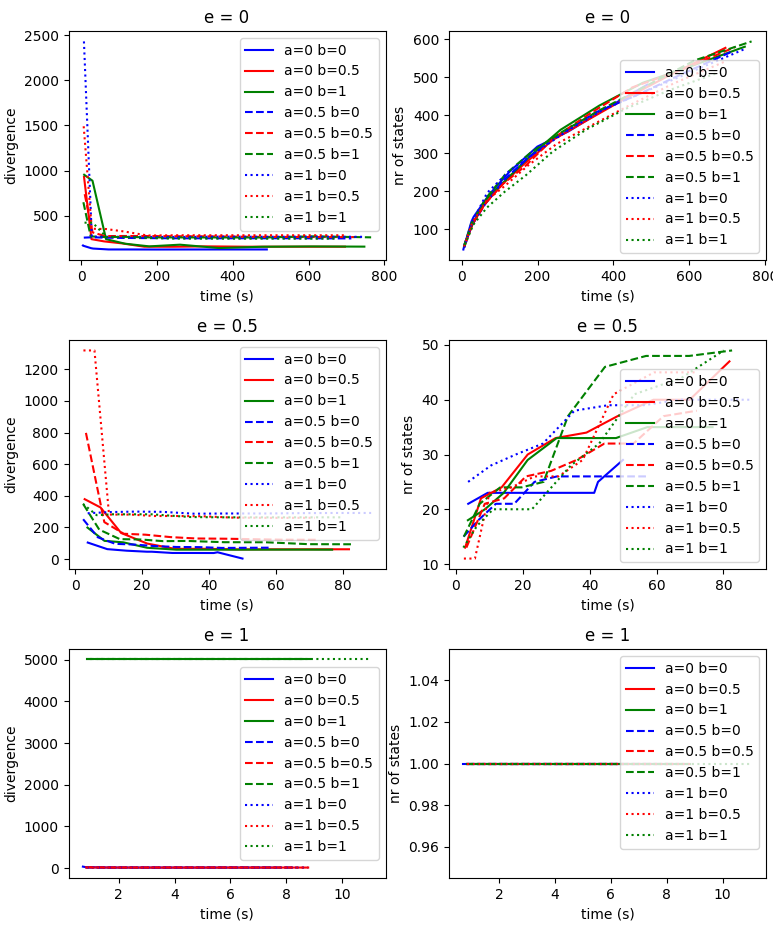}
\caption{Experiments with $5\times5$ building. $a$, $b$, and $e$ stand for $\alpha$, $\beta$, and $\epsilon$, respectively \label{fig:big_experiments}.}
\vspace*{-12pt}
\end{figure}
  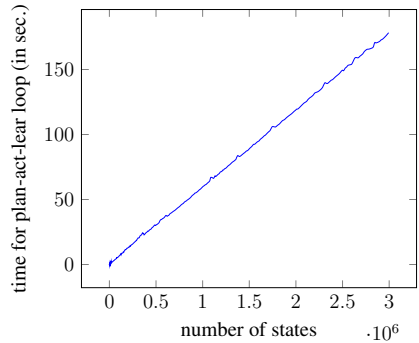
\begin{figure}[t]
    \begin{center}
  \begin{tikzpicture}[scale=.65]
    \begin{axis}[
        xlabel=number of states,
        ylabel=time for plan-act-lear loop (in sec.)]
    \addplot[smooth,blue] plot coordinates {
(2, 0.0067489147)
(4, 0.003376219)
(6, 0.004562974)
(8, 0.0044480562)
(16, 0.004762888)
(32, 0.0075199604)
(40, 0.0070310831)
(48, 0.0057038731)
(56, 0.007139047)
(84, 0.0132210255)
(96, 0.010684967)
(128, 0.0111549695)
(144, 0.0203340054)
(180, 0.0241651535)
(200, 0.0251228809)
(220, 0.0193300247)
(264, 0.0359051228)
(288, 0.0356428623)
(312, 0.0372138023)
(364, 0.0309046507)
(392, 0.0460929871)
(448, 0.0348215103)
(504, 0.0594797134)
(560, 0.0574133396)
(600, 0.0220573213)
(660, 0.0668759346)
(720, 0.0721042156)
(768, 0.0712447166)
(1152, 0.1102471352)
(1224, 0.1147489548)
(1326, 0.0576342344)
(1428, 0.0898090601)
(1512, 0.1519348621)
(1596, 0.0747615695)
(1710, 0.0614803235)
(1824, 0.2179298401)
(1938, 0.0715493858)
(2052, 0.0671575069)
(2160, 0.1916959286)
(2280, 0.1957149506)
(3040, 0.2573492527)
(3200, 0.2632546425)
(4000, 0.3237700462)
(4800, 0.3798379898)
(5040, 0.4114100933)
(5280, 0.4449241161)
(5520, 0.2981020212)
(5760, 0.5135958195)
(6720, 0.5853271484)
(7056, 0.5886199474)
(8064, 0.6604187489)
(8400, 0.6886930466)
(9450, 0.3344310919)
(9828, 0.9019179344)
(10206, 0.6259690523)
(10584, 0.3970844746)
(11088, 1.1125130653)
(11484, 1.2392919064)
(11880, 1.405369997)
(12276, 0.7307783763)
(13640, 1.7276198864)
(14080, 0.7337628433)
(14720, 0.8142523368)
(15180, 1.6492400169)
(15640, 1.7121562958)
(16100, 1.3559203148)
(16800, 1.2501752377)
(18480, 2.018362999)
(19008, 1.3534078598)
(19800, 2.049767971)
(21600, 2.240221262)
(22464, 2.4098176956)
(24336, 1.5433511734)
(25272, 1.583039999)
(27216, 1.7113831043)
(28224, 1.7573900223)
(30240, 1.8922529221)
(32256, 2.0857527256)
(34272, 2.1770093441)
(36288, 2.3112099171)
(37296, 2.3997149467)
(38304, 2.4411950111)
(39312, 2.4905130863)
(41496, 2.617787838)
(42560, 2.7359988689)
(43624, 2.7509922981)
(45920, 2.9472439289)
(47560, 2.9976990223)
(48720, 2.9895241261)
(51156, 3.1724097729)
(53592, 3.3031480312)
(54868, 3.4362411499)
(57362, 3.6587417126)
(58696, 3.7409501076)
(60030, 3.7589008808)
(62640, 3.93475914)
(65250, 4.0865869522)
(67860, 4.1835608482)
(70470, 4.3430252075)
(72036, 4.4319329262)
(74704, 4.7580301762)
(76328, 4.7667310238)
(79054, 5.4087529182)
(80736, 5.3290989399)
(82418, 5.3824810982)
(84100, 5.2663059235)
(85782, 5.2558760643)
(88740, 5.4855699539)
(90480, 5.6401469707)
(93600, 5.704128027)
(96720, 5.9080638885)
(99840, 6.0985851288)
(102960, 6.4040338993)
(104940, 6.6341979504)
(108120, 6.8298249245)
(111724, 7.0605511665)
(115328, 7.419809103)
(118932, 7.6241900921)
(122430, 8.0517537594)
(124740, 7.5704772472)
(127050, 8.3644688129)
(129360, 8.4612140656)
(133280, 8.7122900486)
(137200, 8.7563347816)
(141120, 8.7268249989)
(145040, 8.9900772572)
(147630, 9.5994150639)
(150220, 9.6282639503)
(154280, 9.4534687996)
(158688, 10.2845261097)
(163096, 10.7934279442)
(165908, 11.0059988499)
(170274, 11.5315151215)
(174640, 11.4001851082)
(179360, 11.6834368706)
(184080, 12.0754170418)
(187200, 12.4731268883)
(191880, 12.4784193039)
(196800, 12.7452628613)
(200080, 12.8815348148)
(203360, 13.238117218)
(206640, 13.3985657692)
(209920, 13.2665390968)
(215040, 13.5740180016)
(218400, 13.8671340942)
(221760, 14.4222540855)
(225120, 14.1254336834)
(230748, 14.6462211609)
(234192, 15.0263609886)
(237636, 15.1281909943)
(243294, 15.4521911144)
(246820, 15.5614099503)
(252840, 15.9923567772)
(258720, 16.4510452747)
(264600, 16.8006279469)
(270900, 17.3082039356)
(277200, 17.6330771446)
(283360, 18.1298129559)
(287408, 18.4245560169)
(291456, 18.4273648262)
(295504, 19.2641301155)
(301928, 19.7505099773)
(308352, 20.4928929806)
(315360, 20.410351038)
(322368, 20.8786480427)
(326784, 21.8218610287)
(333888, 22.2834308147)
(338400, 22.5960412025)
(345600, 22.8637137413)
(352800, 24.0919251442)
(357504, 24.4245181084)
(362208, 23.8944597244)
(366912, 23.8221073151)
(371616, 22.766302824)
(376320, 23.1194930077)
(384000, 23.6576800346)
(391680, 24.3936240673)
(399360, 24.7449700832)
(407680, 25.3131537437)
(412776, 25.7215249538)
(417872, 25.5914392471)
(422968, 25.9493298531)
(431102, 26.6355571747)
(436296, 26.8618757725)
(444528, 27.1683630943)
(449820, 27.523567915)
(455112, 28.1612970829)
(463540, 28.5742139816)
(468930, 29.0427041054)
(478500, 29.7999749184)
(488070, 30.434926033)
(493680, 30.0378491879)
(502656, 30.4535849094)
(508368, 30.7800588608)
(517446, 31.1200489998)
(523260, 31.5796098709)
(532440, 32.1562922001)
(538356, 33.1775429249)
(548912, 34.4012451172)
(558376, 34.5776209831)
(564512, 34.9593999386)
(570648, 35.3220880032)
(580320, 35.6756949425)
(589992, 36.3177821636)
(596336, 36.9358139038)
(602680, 37.518903017)
(609024, 37.7333619595)
(619008, 37.1076178551)
(628992, 37.59922719)
(635544, 37.7491416931)
(642096, 38.3408093452)
(652288, 39.2471618652)
(662480, 39.6109850407)
(672672, 40.1911058426)
(682864, 40.7430891991)
(689832, 41.0949409008)
(700128, 41.5502710342)
(710424, 42.1555309296)
(724086, 43.0329830647)
(734580, 43.674957037)
(742000, 44.0504920483)
(752600, 44.6265878677)
(763200, 45.3909208775)
(770832, 45.9417703152)
(785376, 46.7034928799)
(793152, 47.0339269638)
(800928, 47.7287139893)
(812052, 48.4212598801)
(827090, 49.0077362061)
(835120, 49.620994091)
(843150, 49.9250628948)
(858480, 50.7853097916)
(866656, 51.276373148)
(874832, 51.8978641033)
(883008, 52.6403849125)
(891184, 53.2266819477)
(907098, 54.0899062157)
(915420, 54.8437209129)
(923742, 55.0315380096)
(936396, 55.5852379799)
(944832, 56.0783548355)
(953268, 56.6197252274)
(966150, 57.5517580509)
(974700, 57.9059479237)
(987696, 58.8314900398)
(1005024, 60.0806016922)
(1018248, 60.7036690712)
(1027180, 61.2934901714)
(1036112, 61.5354897976)
(1049568, 62.4769310951)
(1058616, 63.059014082)
(1076868, 64.383163929)
(1086072, 66.9460458755)
(1104480, 66.7878582478)
(1113840, 66.2605457306)
(1123200, 66.7849190235)
(1132560, 68.201690197)
(1141920, 67.7375628948)
(1160952, 69.0527980328)
(1170468, 69.566519022)
(1179984, 70.1197638512)
(1189500, 71.2019851208)
(1204750, 71.5152671337)
(1214388, 72.574244976)
(1224026, 72.6780838966)
(1233664, 73.5488550663)
(1249280, 74.1220390797)
(1269760, 75.6274659634)
(1279680, 76.2692258358)
(1289600, 76.7079801559)
(1310400, 78.1776578426)
(1320480, 78.8226811886)
(1336986, 79.6090359688)
(1347192, 80.2723069191)
(1357398, 80.9705460072)
(1374156, 83.7214109898)
(1395968, 83.0790131092)
(1417780, 84.1619238853)
(1439592, 85.4907710552)
(1450416, 86.0146780014)
(1461240, 86.7724628448)
(1472064, 87.4099342823)
(1482888, 87.6541507244)
(1505356, 89.416282177)
(1523714, 90.5838007927)
(1534836, 91.504420042)
(1545958, 92.4505901337)
(1557080, 92.4675278664)
(1568202, 93.1511631012)
(1587096, 94.3725450039)
(1605990, 95.1598689556)
(1617380, 96.2654788494)
(1641520, 97.6442670822)
(1660832, 99.0410621166)
(1672528, 99.4979150295)
(1691976, 100.9939959049)
(1711424, 102.3206760883)
(1723392, 103.0949509144)
(1742976, 105.7566480637)
(1755080, 106.1413071156)
(1780890, 105.7638418674)
(1800900, 106.9808630943)
(1820910, 107.9669167995)
(1833468, 108.9120020866)
(1853616, 110.5015721321)
(1866312, 111.0440819263)
(1879008, 111.5390560627)
(1899432, 112.9739968777)
(1926960, 114.557945013)
(1954488, 116.0113868713)
(1982016, 117.8937621117)
(1995408, 118.6362831593)
(2008800, 119.546202898)
(2022192, 119.6734981537)
(2050278, 121.6077187061)
(2078364, 123.6441862583)
(2092128, 124.4457507133)
(2114624, 125.4744150639)
(2128536, 126.8433520794)
(2151180, 128.3532669544)
(2165240, 129.0013079643)
(2188032, 130.4993929863)
(2202240, 131.9331991673)
(2232000, 132.7949278355)
(2255250, 133.9410102367)
(2285320, 136.1493577957)
(2308880, 139.6985452175)
(2332440, 139.2168467045)
(2347488, 139.7243990898)
(2378376, 141.7711040974)
(2393622, 142.2280330658)
(2424708, 144.3945960999)
(2455794, 146.2203588486)
(2471436, 147.3158609867)
(2496400, 149.384403944)
(2512200, 149.1159350872)
(2528000, 151.0094020367)
(2543800, 151.9083418846)
(2559600, 153.2584750652)
(2585196, 153.6955718994)
(2601154, 154.4232599735)
(2634080, 158.9747350216)
(2667006, 158.9796750546)
(2683368, 160.0517511368)
(2716496, 161.983112812)
(2733060, 163.8062539101)
(2760120, 165.4422709942)
(2776848, 165.6322653294)
(2793576, 165.9032287598)
(2827644, 167.1105868816)
(2844576, 170.6325449944)
(2861508, 170.4805603027)
(2895984, 171.5433828831)
(2913120, 172.749281168)
(2930256, 173.3917768002)
(2965140, 175.8460478783)
(2994210, 178.3089110851)
      };
    \end{axis}
  \end{tikzpicture}
\end{center}
\caption{\label{fig:plot scalability} Time for execution of the 
    plan-act-learn loop depending on the size of $S$.\vspace*{-12pt}}
\end{figure}  

\section{Related Work}
\label{sec:rw}
Our approach shares some similarities with the work on planning by
model based reinforcement learning (RL)
\cite{sutton98}. 
In \cite{yang2018,lyu2019}, symbolic planning is based on the action
language ${\mathcal BC}$ in a hierarchical (deep) reinforcement learning
setting. 
In \cite{parr97} hierarchical abstract machines impose
constraints on reinforcement learning.  \cite{ryan2002} combines
symbolic planning techniques with reinforcement learning.  In
\cite{leonetti2016} plans are generated by answer set programming, and
reinforcement learning allows adaptation to a changing environment.
\cite{henaff2017ArXiv} proposes 
model based planning to learn the planning domain directly from
execution traces.
All the works mentioned above assume that
the set of states and the correspondence between continuous data from
perceptions and states are fixed a priori and remains unchanged.  

The two closest approaches to our proposal are Causal InfoGAN
\cite{kurutach2018} and LatPlan \cite{asai2018}. 
Causal InfoGan learns discrete or continuous models from high
dimensional sequential observations.
Latent (abstract) representation of states, and mapping between
states and observations are learned by maximizing the mutual
information between the generated observations and the transition in
the planning model. This approach fixes a priori the size of the
discrete domain model, and performs the learning off line. Differently
from our approach their goal is to generate an execution trace in the
high dimensional space.

  LatPlans takes in input pairs of high dimensional
  raw data (e.g., images) corresponding to transitions. It takes an
  offline approach. In a first phase, a State Autoencoder learns a
  mapping between raw data and abstract states, represented as vectors
  of binary state variables.  In the second phase, LatPlan learns a
  transition function from the state pairs obtained by applying the
  mapping learned in the first phase to the training pairs. Planning
  is finally applied to the learned model. From the one hand, LatPlan
  has been shown experimentally to work with high dimensional data
  like images. From the other hand, our approach is online: \PAL\
  indeed interleaves learning, planning, and acting phases at run time. This
  allows \PAL\ to deal with a dynamic environment. Moreover,
  \PAL\ doesn't fix the state space a priori, while in LatPlan
  one has to fix the dimension of the state embedding.

A complementary approach is pursued in works that plan directly in a
continuous space, e.g., \cite{abbeel2006,mnih2015,coeeyes2018}.
In this way there is no need of a perception function, since there is no abstract discrete model of the
world. Such approaches are very suited to address some tasks, e.g., moving a
robot arm to a desired position or performing some manipulations. 
However, we believe that, in several situations, it is
conceptually appropriate and practically efficient to perform planning
in an abstract discrete state space. 

Several approaches in robotics (see Sect.7 of \cite{ingrand2017} for a survey) deal with the problem
of planning in and learning the environment in which they operate, and
they have to deal with the robot ending up in unknown and unexpected
states of the world.  Some of them make use of an abstract model of
the world.
However, none of these works provide a formalization of the mapping
and of the learning mechanism as we provide in this paper.
%
%
%
Works on domain model acquisition focus on the different problem of
learning action schema, see,
e.g. \cite{gregory2016,cresswell2013}.

\section{Conclusions and Future Work}
\label{sec:conclusion}

We have provided a formal framework for the  online 
construction of an abstract planning domain by learning new states
and the mapping between states and continuous perceptions. 
Our experiments show convergence to coherent models. In the future, we will provide a formal proof
of convergence under some specific assumptions (i.e., the convergence of formula
\eqref{eq:coherence-approx} to a finite value). 
We share with several works the intuition that planning at the
abstract discrete level might be convenient in some application
domains. In the future, we plan to compare our approach with
approaches that do planning in a continuous space. 
In this paper we focus on fully observable deterministic domains. We
plan to extend our work to (partial observable) nondeterministic and
probabilistic domains, e.g., by learning probability distributions on
$\gamma$.%
\footnote{We assume full observability,
    since we select only one state with max-likelihood, rather than a
    set of states after action execution.  In the work on Partial
    Observability in Non-deterministic Domains (POND), and in
    stochastic domains (POMDP), after action execution, the agent
    knows to be in a set of states or in a probability distribution of
    over the set of states, see, e.g., \cite{geffner2013}.}
Moreover, we plan to integrate in our framework a state-of-the-art
on-line planner, and to run experiments on more complex and realistic
domains. 
Although much work is required to determine the
  applicability and scalability of this approach, we believe this work
  is an important first step in bridging the gap between online planning in a
  discrete abstract model and perceptions in a continuous changing space.

\bibliographystyle{plain}
\bibliography{biblio}

\begin{thebibliography}{10}

\bibitem{abbeel2006}
Pieter Abbeel, Morgan Quigley, and Andrew~Y. Ng.
\newblock Using inaccurate models in reinforcement learning.
\newblock In {\em Machine Learning, Proceedings of the Twenty-Third
  International Conference {(ICML} 2006), Pittsburgh, Pennsylvania, USA, June
  25-29, 2006}, pages 1--8, 2006.

\bibitem{asai2018}
Masataro Asai and Alex Fukunaga.
\newblock Classical planning in deep latent space: Bridging the
  subsymbolic-symbolic boundary.
\newblock In {\em Proceedings of the Thirty-Second {AAAI} Conference on
  Artificial Intelligence, (AAAI-18)}, pages 6094--6101, 2018.

\bibitem{bishop2006}
Christopher~M. Bishop.
\newblock {\em Pattern Recognition and Machine Learning (Information Science
  and Statistics)}.
\newblock Springer-Verlag, Berlin, Heidelberg, 2006.

\bibitem{coeeyes2018}
John~D. Co{-}Reyes, Yuxuan Liu, Abhishek Gupta, Benjamin Eysenbach, Pieter
  Abbeel, and Sergey Levine.
\newblock Self-consistent trajectory autoencoder: Hierarchical reinforcement
  learning with trajectory embeddings.
\newblock In {\em Proceedings of the 35th International Conference on Machine
  Learning, {ICML} 2018, Stockholmsm{\"{a}}ssan, Stockholm, Sweden, July 10-15,
  2018}, pages 1008--1017, 2018.

\bibitem{cresswell2013}
Stephen Cresswell, Thomas~Leo McCluskey, and Margaret~Mary West.
\newblock Acquiring planning domain models using \emph{LOCM}.
\newblock {\em Knowledge Eng. Review}, 28(2):195--213, 2013.

\bibitem{geffner2013}
Hector Geffner and Blai Bonet.
\newblock {\em A Concise Introduction to Models and Methods for Automated
  Planning}.
\newblock Synthesis Lectures on Artificial Intelligence and Machine Learning.
  Morgan {\&} Claypool Publishers, 2013.

\bibitem{ghallab2004}
Malik Ghallab, Dana~S. Nau, and Paolo Traverso.
\newblock {\em Automated Planning - Theory and Practice}.
\newblock Elsevier, 2004.

\bibitem{ghallab2016}
Malik Ghallab, Dana~S. Nau, and Paolo Traverso.
\newblock {\em Automated Planning and Acting}.
\newblock Cambridge University Press, 2016.

\bibitem{gregory2016}
Peter Gregory and Stephen Cresswell.
\newblock Domain model acquisition in the presence of static relations in the
  {LOP} system.
\newblock In {\em Proceedings of the Twenty-Fifth International Joint
  Conference on Artificial Intelligence, {IJCAI} 2016, New York, NY, USA, 9-15
  July 2016}, pages 4160--4164, 2016.

\bibitem{henaff2017ArXiv}
M.~{Henaff}, W.~F. {Whitney}, and Y.~{LeCun}.
\newblock {Model-Based Planning with Discrete and Continuous Actions}.
\newblock {\em ArXiv e-prints}, 2017.

\bibitem{ingrand2017}
F{\'{e}}lix Ingrand and Malik Ghallab.
\newblock Deliberation for autonomous robots: {A} survey.
\newblock {\em Artif. Intell.}, 247:10--44, 2017.

\bibitem{kurutach2018}
Hanard Kurutach, Aviv Tamar, Ge~Yang, Stuart Russell, and Pieter Abbeel.
\newblock Learning plannable representations with causal infogan.
\newblock In {\em NIPS}, 2018.

\bibitem{leonetti2016}
Matteo Leonetti, Luca Iocchi, and Peter Stone.
\newblock A synthesis of automated planning and reinforcement learning for
  efficient, robust decision-making.
\newblock {\em Artif. Intell.}, 241:103--130, 2016.

\bibitem{lyu2019}
Daoming Lyu, Fangkai Yang, Bo~Liu, and Steven Gustafson.
\newblock {SDRL}: Interpretable and data-efficient deep reinforcement learning
  leveraging symbolic planning.
\newblock In {\em AAAI 2019}, 2019.
\newblock (To appear).

\bibitem{mnih2015}
Volodymyr Mnih, Koray Kavukcuoglu, David Silver, Andrei~A. Rusu, Joel Veness,
  Marc~G. Bellemare, Alex Graves, Martin~A. Riedmiller, Andreas Fidjeland,
  Georg Ostrovski, Stig Petersen, Charles Beattie, Amir Sadik, Ioannis
  Antonoglou, Helen King, Dharshan Kumaran, Daan Wierstra, Shane Legg, and
  Demis Hassabis.
\newblock Human-level control through deep reinforcement learning.
\newblock {\em Nature}, 518(7540):529--533, 2015.

\bibitem{parr97}
Ronald Parr and Stuart~J. Russell.
\newblock Reinforcement learning with hierarchies of machines.
\newblock In {\em Advances in Neural Information Processing Systems 10, {[NIPS}
  Conference, Denver, Colorado, USA, 1997]}, pages 1043--1049, 1997.

\bibitem{ryan2002}
Malcolm R.~K. Ryan.
\newblock Using abstract models of behaviours to automatically generate
  reinforcement learning hierarchies.
\newblock In {\em Machine Learning, Proceedings of the Nineteenth International
  Conference {(ICML} 2002), University of New South Wales, Sydney, Australia,
  July 8-12, 2002}, pages 522--529, 2002.

\bibitem{sutton90}
Richard~S. Sutton.
\newblock Integrated architectures for learning, planning, and reacting based
  on approximating dynamic programming.
\newblock In {\em Machine Learning, Proceedings of the Seventh International
  Conference on Machine Learning, Austin, Texas, USA, June 21-23, 1990}, pages
  216--224, 1990.

\bibitem{sutton98}
Richard~S. Sutton and Andrew~G. Barto.
\newblock {\em Reinforcement learning - an introduction}.
\newblock Adaptive computation and machine learning. {MIT} Press, 1998.

\bibitem{yang2018}
Fangkai Yang, Daoming Lyu, Bo~Liu, and Steven Gustafson.
\newblock {PEORL:} integrating symbolic planning and hierarchical reinforcement
  learning for robust decision-making.
\newblock In {\em Proceedings of the Twenty-Seventh International Joint
  Conference on Artificial Intelligence, {IJCAI} 2018, July 13-19, 2018,
  Stockholm, Sweden.}, pages 4860--4866, 2018.

\end{thebibliography}
\section*{Acknowledgments}
  We thank the anonymous reviewers of a previous submission.
  Their comments allowed us to significantly improve our paper.
\end{document}